\documentclass{article}
\usepackage{ijcai26}

\usepackage{times}
\usepackage{soul}
\usepackage{url}
\usepackage[hidelinks]{hyperref}
\usepackage[utf8]{inputenc}
\usepackage[small]{caption}
\usepackage{graphicx}
\usepackage{amsmath}
\usepackage{amsthm}
\usepackage{booktabs}
\usepackage{algorithm}
\usepackage[switch]{lineno}
\usepackage{xcolor}
\usepackage{multirow}
\usepackage{tabularx}

\usepackage[table]{xcolor}
\usepackage{amsfonts}
\usepackage{algpseudocode}
\usepackage{pifont}
\usepackage{makecell}
\usepackage{eso-pic}
\usepackage{array}

\newcolumntype{Y}{>{\centering\arraybackslash}X}
\newcommand{\cmark}{\ding{51}}
\newcommand{\xmark}{\ding{55}}

\title{Learning with Semantic Priors: Stabilizing Point-Supervised Infrared Small Target Detection via Hierarchical Knowledge Distillation}

\author{
Yuanhang Yao$^1$\and
Ping Qian$^1$\and
Zhu Liu$^1$\and
Long Ma$^1$\and
Weimin Wang$^1$\thanks{Corresponding author.}\\
\affiliations
$^1$School of Software Technology, Dalian University of Technology, China\\
\emails
yuanhangyao2027@gmail.com,
wangweimin@dlut.edu.cn
}

\begin{document}

\pagestyle{plain}
\thispagestyle{plain}
\AddToShipoutPictureFG*{
  \AtTextLowerLeft{
    \raisebox{-20pt}{
      \makebox[0pt][l]{\normalfont\footnotesize Accepted to IJCAI-ECAI 2026.}
    }
  }
}

\maketitle

\begin{abstract}

Single-frame Infrared Small Target Detection (ISTD) aims to localize weak targets under heavy background clutter, yet dense pixel-wise annotations are expensive. Point supervision with online label evolution reduces annotation cost; however, lightweight CNN detectors often lack sufficient semantics, leading to noisy pseudo-masks and unstable optimization. To address this, we propose a hierarchical VFM-driven knowledge distillation framework that uses a frozen Vision Foundation Model (VFM) during training. We formulate point-supervised learning as a bilevel optimization process: the inner loop adapts a VFM-embedded teacher on reweighted training samples, while the outer loop transfers validation-guided knowledge to a lightweight student to mitigate pseudo-label noise and training-set bias. We further introduce Semantic-Conditioned Affine Modulation (SCAM) to inject VFM semantics into CNN features at multiple layers. In addition, a dynamic collaborative learning strategy with cluster-level sample reweighting enhances robustness to imperfect pseudo-masks. Experiments on diverse challenging cases across multiple ISTD backbones demonstrate consistent improvements in detection accuracy and training stability. Our code is available at \url{https://github.com/yuanhang-yao/semantic-prior}.

\end{abstract}

\begin{figure*}[!htb]
	\centering
	\setlength{\tabcolsep}{1pt} 
	\begin{tabular}{c}		
		\includegraphics[width=0.99\textwidth]{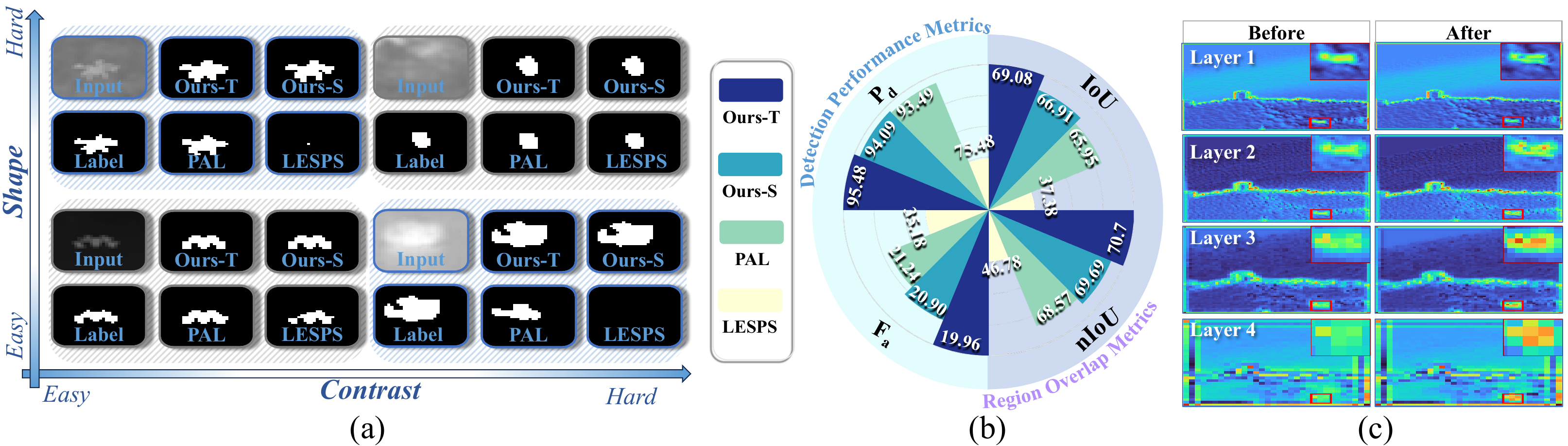}\\
	\end{tabular}
	\caption{
Efficiency overview.
(a) Qualitative predictions on four characteristic test subsets (Salient, Filamentary, Faint, Camouflaged), spanning easy-to-hard contrast and shape.
(b) Quantitative results on SCTransNet, where we achieve state-of-the-art on both region overlap and detection performance. 
(c) VFM-guided modulation sharpens target-aware features while suppressing background clutter.
}
\label{fig:head}
\end{figure*}

\section{Introduction}
Single-frame Infrared Small Target Detection (ISTD) is a vital technique in infrared search and tracking systems, owing to its superior capabilities of passive detection and all-weather operation. The goal is to detect faint signals  within intricate settings characterized by severe background clutter and real-world degradations~\cite{liu2025deal,wang2024mergenet,wang2023snow}, with broad applications in early warning and maritime search and rescue. Despite the success of fully supervised methods, their scalability is severely limited by the prohibitive cost of annotation. Consequently, point supervision, which requires only one coordinate annotation per target, has emerged as a challenging direction.

In the early stage, methods were dominated by hand-crafted priors. Representative techniques, such as filter-based methods (e.g., Top-Hat~\cite{bai2010analysis}) and Human Visual System (HVS)-based methods (e.g., LCM~\cite{chen2013local}), utilize local contrast mechanisms to suppress background clutter. However, these methods rely on the adjustments of hyper-parameters and specific assumptions, limiting generalization in complex scenarios~\cite{liu2023enhancing}. Recently, with the paradigm shift toward data-driven semantic segmentation, deep learning-based methods~\cite{wang20253d,wang2026swg} have dominated the ISTD field. To preserve the signatures of tiny targets from corrupted features, diverse mechanisms are proposed: multi-scale feature fusion (e.g., UIUNet~\cite{wu2022uiu}) effectively integrates high-level semantics with low-level spatial details; contextual modulation schemes (e.g., ACM~\cite{dai2021asymmetric}) utilize local contrast and attention to highlight dim targets against complex backgrounds; and dense nested interactions (e.g., DNANet~\cite{li2022dense}) enable repetitive feature enhancement to maintain signal strength. However, fully supervised annotation is labor-intensive, limiting scalability.

To address the annotation bottleneck, point-supervised methods have recently emerged as a promising direction. LESPS~\cite{Ying_2023_CVPR} pioneered this paradigm, introducing a label evolution strategy to iteratively update pseudo-masks from point signals. But the unconstrained evolution in LESPS often suffers from optimization instability. To mitigate this, PAL~\cite{yu2025easy} proposed an easy-to-hard active learning curriculum to stabilize the training trajectory. Although these methods are effective, a fundamental issue remains unaddressed: the semantic deficiency in lightweight ISTD networks. In the online label evolution framework, the generation of pseudo-labels heavily depends on the feature extraction capability. However, shallow features lack high-level semantic discrimination, making it difficult to distinguish dim targets from high-frequency noise. This semantic gap leads to the generation of noisy pseudo-labels, causing training oscillation and performance degradation.

The emergence of Vision Foundation Model (VFM)~\cite{radford2021learning,kirillov2023segment,oquab2023dinov2} has catalyzed a paradigm shift in ISTD, formally characterized as a foundation-driven paradigm. Capitalizing on this, SAM-based adaptations like IRSAM~\cite{zhang2024irsam} exploit tailored prompt encoders and self-prompting mechanisms to transfer open-world segmentation capabilities to the infrared domain. Concurrently, vision-language approaches such as SAIST~\cite{zhang2025saist} leverage CLIP to align textual semantics with visual features for enhanced discrimination ability. However, despite their superior representation power, these methods inevitably incur a prohibitive inference burden due to their reliance on heavy, over-parameterized backbones. This heavy-encoder design results in redundant computation and high latency, fundamentally hindering their deployment on resource-constrained edge devices for real-time use.

Driven by these observations, a critical research question arises: How can we leverage the robust semantic priors of VFMs to rectify the feature insufficiency of point supervision, while  preserving the inference efficiency? This objective presents a dual challenge. On one hand, the network can effectively distill the rich semantic representation and generalization capability from frozen VFMs to bridge the semantic gap  of point-supervised networks, thereby stabilizing the label evolution process. On the other hand, it can achieve a complete decoupling of the heavy parameters from the inference pipeline. Furthermore, direct knowledge distillation straightforwardly minimizes the discrepancy between  students and teachers on the training set. However, we argue that this naive alignment creates a generalization bottleneck: the student tends to overfit to the training domain bias.

To address the instability and generalization bottlenecks in online point supervision, we propose a  hierarchical VFM-driven knowledge distillation framework. By reformulating the learning paradigm as a hierarchical optimization process, we enable the model to dynamically optimize for representation fitting (lower level) and generalization feedback (upper level). In detail, through explicitly optimizing for performance on an unseen validation distribution in the upper level, we generate VFM-guided feedback that prevents the student network from overfitting to the bias of  training set. Furthermore, we design the semantic-conditioned affine modulation to inject robust object-centric semantics from frozen VFMs into the student network, rectifying the semantic deficiency of shallow features. Moreover, to combat overfitting risks, we introduce a dynamic collaborative learning strategy with cluster-level sample reweighting. This mechanism dynamically down-weights outliers while emphasizing informative hard samples, significantly enhancing both the robustness of label evolution and the detection performance in complex scenarios. Our contributions can be summarized as:

\begin{itemize}
\item We propose a VFM-driven knowledge distillation framework for point-supervised ISTD by reformulating online point supervision as a bilevel optimization problem. We prioritize validation-based generalization feedback, improving training stability under evolving pseudo-masks.

\item To bridge the semantic gap between VFM priors and the fine-grained semantics required by tiny-target localization, we design the semantic-conditioned affine modulation, a lightweight interface that injects VFM semantics into the shallow ISTD network.

\item A dynamic collaborative learning strategy is introduced to enhance the robustness of label evolution under noisy pseudo-masks and severe data imbalance. By cluster-level sample reweighting, it automatically down-weights noisy pseudo-labels without manual tuning.

\item Experiments on SIRST benchmarks demonstrate consistent improvements across multiple backbones, while keeping the student network efficient at inference.
\end{itemize}

\begin{figure*}[!htb]
\centering
\setlength{\tabcolsep}{1pt} 
\begin{tabular}{c}		
\includegraphics[width=0.99\textwidth]{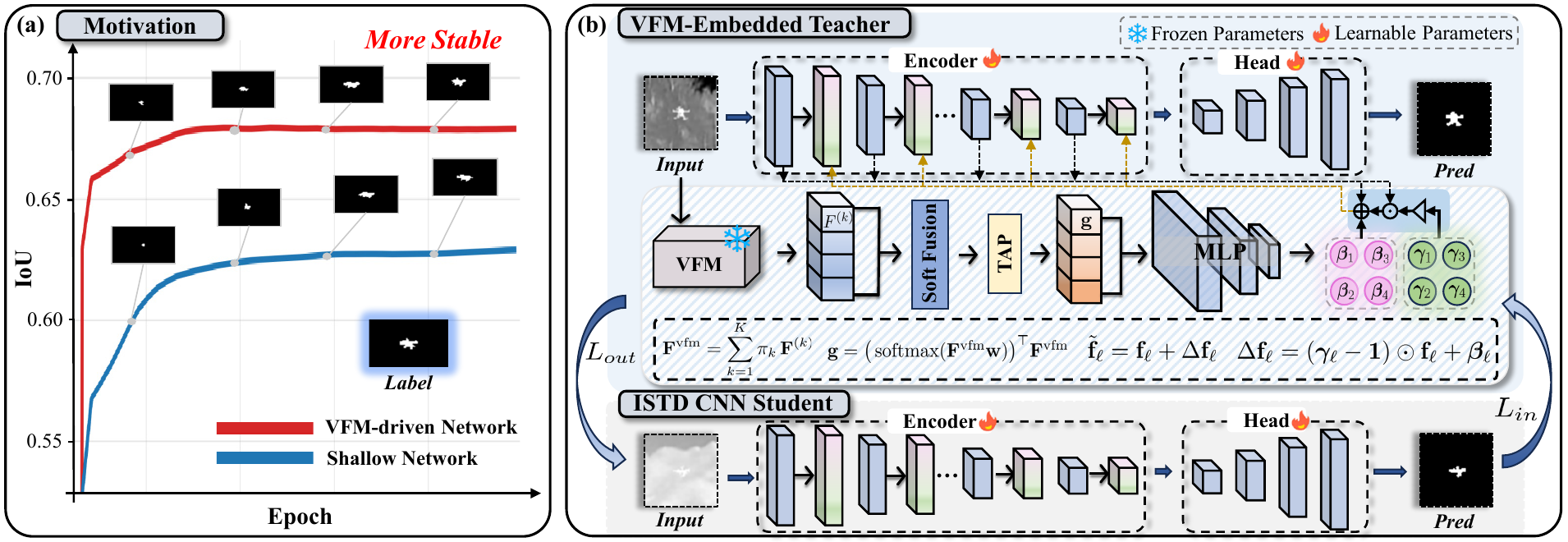}\\
\end{tabular}
\caption{
(a) Motivation of introducing VFM semantics to stabilize online label evolution in point-supervised ISTD. (b) The proposed hierarchical VFM-driven distillation framework with a VFM-embedded teacher, SCAM-based modulation, and an ISTD CNN student.
}
\label{fig:process}
\end{figure*}

\section{The Proposed Method}

\subsection{VFM-driven Distillation Framework}
\label{sec:vfm_distill}

As mentioned above, pseudo-masks are inevitably noisy in the early phase of online evolution, and their evolution quality is strongly coupled with the feature extractor.
Lightweight CNNs may entangle weak targets with high-frequency clutter, causing unstable mask updates and fragile representations.
VFMs possess stronger target-centered semantics and better cross-domain robustness~\cite{radford2021learning}, making them suitable for addressing the semantic deficiency of point supervision.
However, directly deploying VFMs is often impractical due to high inference costs and a mismatch with the semantic granularity of small targets.
A naive alternative is to distill students on the training set, but this direct alignment may become a generalization bottleneck: under the induction of evolving pseudo-masks, students tend to imitate the teacher on the training distribution, overfitting to dataset-specific textures instead of inheriting transferable semantics.

Therefore, motivated by recent advances in bilevel optimization~\cite{liu2023bi,liu2024moreau}, we propose a hierarchical VFM knowledge distillation framework that uses a frozen VFM only during training and explicitly optimizes for generalization capabilities, which can be formulated as: 
\begin{equation}
\begin{aligned}
\min_{\theta,\alpha}\quad &
\mathbb{E}_{(x,\tilde{y})\sim \mathcal{D}_{\mathrm{val}}}
\Big[\, L_{\mathrm{out}}(x,\tilde{y};\theta,\alpha,\phi^\star)\,\Big] \\
\text{s.t.}\quad &
\phi^\star
=\arg\min_{\phi}\;
\mathbb{E}_{(x,\tilde{y})\sim \mathcal{D}_{\mathrm{tr}}}
\Big[\, L_{\mathrm{in}}(x,\tilde{y};\theta,\alpha,\phi)\,\Big].
\end{aligned}
\end{equation}
Specifically, we introduce the base ISTD network, denoted as $S_{\theta}$, and a VFM-embedded teacher network, denoted as $T_{\theta,\phi}$, where $\phi$ denotes the teacher-side adaptation parameters, and $\alpha$ denotes the parameters of the sample reweighting function $w_\alpha(\cdot)$. The teacher shares the student weights $\theta$ but additionally equips a SCAM modulator (reported in Sec.~\ref{sec:scam}) parameterized by $\phi$, which injects semantic information into student layers from the VFM.
The inner objective $L_{\mathrm{in}}$ is to fit a representation on weighted training data and update teacher-side adaptation parameters $\phi$, enabling it to generate more reliable soft labels and stabilize label evolution:
\begin{equation}
\begin{aligned}
L_{\mathrm{in}}
&= w_{\alpha}(x)\Big(
L_{\mathrm{task}}(p_t,\tilde{y})
+ \lambda_{\mathrm{in}}\,L_{\mathrm{kd}}(p_t,p_s)
\Big) 
 + \lambda_{\mathrm{gate}}\,\mathcal{R}_{\mathrm{gate}},
\end{aligned}
\end{equation}
where $\mathcal{R}_{\mathrm{gate}}=\sum_{\ell}\lVert \mathbf{u}_{\ell}\rVert_{1}$ is set to encourage sparsity in the gating variables of semantic modulation $\mathbf{u}_\ell$, so that the teacher selectively injects VFM semantics only when beneficial, yielding a more compressible guidance signal for the CNN student.
Here, $p_t$ and $p_s$ denote the teacher and student predictions, respectively.
We set $L_{\mathrm{task}}(p,\tilde{y})=\mathrm{BCE}(p,\tilde{y})$.
For knowledge distillation, we use the temperature-scaled sigmoid distributions
$q_s=\sigma(z_s/\tau)$ and $q_t=\sigma(z_t/\tau)$ with $\tau=4$, and define
$L_{\mathrm{kd}}$ as the mean-squared error between $q_s$ and $q_t$.
$\lambda_{\mathrm{in}}$ controls the inner KD strength and $\lambda_{\mathrm{gate}}$ weights the sparsity regularizer $\mathcal{R}_{\mathrm{gate}}$.
We set $\lambda_{\mathrm{in}}$ to $0.1$ and $\lambda_{\mathrm{gate}}$ to $5\times10^{-3}$.
The outer objective $L_{\mathrm{out}}$ conducts generalization-driven distillation on validation distributions, jointly optimizing the student and sample weights, so that knowledge transfer is more inclined to improve validation performance rather than merely matching the training set:
\begin{equation}
L_{\mathrm{out}}
= w_{\alpha}(x)\Big(
L_{\mathrm{task}}(p_s,\tilde{y})
+ \lambda_{\mathrm{out}}\,L_{\mathrm{kd}}(p_s,p_t)
\Big),
\end{equation}
where $\lambda_{\mathrm{out}}$ controls the outer KD strength and is set to $1.0$ in all experiments.
We detach the student prediction in the inner-level KD and detach the teacher prediction in the outer-level KD, since the teacher and student share $\theta$.

\subsection{Semantic-Conditioned Affine Modulation}
\label{sec:scam}

The practical bottleneck of introducing a VFM into point-supervised ISTD lies in the semantic granularity mismatch: Transformer token features have strong semantics but coarse spatial resolution, while infrared small targets rely on fragile local cues and multi-scale CNN priors. Existing fusion methods are either very heavy (e.g., cross-attention) or overly coarse (e.g., concatenation / global pooling), which can easily lead to blurring of small targets, unstable gradients under noisy pseudo-masks, and significantly increase training and inference overhead. Therefore, we aim to find a training-phase mechanism to inject VFM semantics into CNN detectors in a lightweight and controllable manner.
We propose Semantic-Conditioned Affine Modulation (SCAM), which converts the frozen VFM features into multi-layer affine parameters for modulating CNN backbone features.
For an input image $x\in\mathbb{R}^{H\times W}$, the frozen VFM outputs token embeddings. We denote the token features from the $k$-th selected transformer block as
$
\mathbf{F}^{(k)}(x)\in\mathbb{R}^{N\times d},
$
where $N$ is the number of tokens, and $d$ is the token embedding dimension.
We then learn a soft fusion across depths from $K$ blocks:
\begin{equation}
\label{eq:vfm_fuse}
\mathbf{F}^{\mathrm{vfm}}=\sum_{k=1}^{K}\pi_k\,\mathbf{F}^{(k)},
\end{equation}
where $\boldsymbol{\pi}=\{\pi_k\}_{k=1}^{K}$ are learnable weights with $\sum \pi_k=1$, and $K$ is set to $12$ to maximize the number of selected layers.

Inspired by DynamicViT~\cite{rao2021dynamicvit}, we introduce Token-aware Attention Pooling (TAP) to aggregate $\mathbf{F}^{\mathrm{vfm}}$ into a global semantic vector $\mathbf{g}$.
Specifically, given $\mathbf{F}^{\mathrm{vfm}}=[t_1,\dots,t_N]\in\mathbb{R}^{N\times d}$, TAP assigns an attention weight to each token via a lightweight scoring head:
\begin{equation}
\label{eq:TAP}
a_j=\frac{\exp(\mathbf{w}^\top t_j)}{\sum_{m=1}^{N}\exp(\mathbf{w}^\top t_m)},\quad
\mathbf{g}=\sum_{j=1}^{N} a_j t_j \in\mathbb{R}^{d},
\end{equation}
where $\mathbf{w}$ is a learnable parameter vector and $a_j$ are normalized attention scores.
Unlike global average pooling, this design better preserves tokens related to the target, facilitating semantic stability for small targets in complex backgrounds.

Subsequently, SCAM maps $\mathbf{g}$ to the affine parameters $(\boldsymbol{\gamma},\boldsymbol{\beta})$ of each layer in the CNN. For the feature $\mathbf{f}_\ell$ of the layer $\ell$, a compact MLP generates $(\boldsymbol{\gamma}_\ell,\boldsymbol{\beta}_\ell)$ with inputs $\mathbf{g}$, where $\boldsymbol{\gamma}_\ell, \boldsymbol{\beta}_\ell \in\mathbb{R}^{C_\ell}$ represent the channel scale and bias. We introduce a channel gating vector $\mathbf{u}_\ell$ to filter $\boldsymbol{\gamma}_\ell$, enabling selective semantic injection and providing an interface for subsequent sparse regularization. To ensure stable training under online pseudo-masking, we adopt a residual form of scale transformation:
$
\boldsymbol{\gamma}_\ell \leftarrow 1+\tanh(\boldsymbol{\gamma}_\ell \odot \mathbf{u}_\ell).
$
Finally, after $\boldsymbol{\gamma}_\ell, \boldsymbol{\beta}_\ell$ are broadcast along spatial dimensions, we modulate the features through an affine transformation in a residual form:
\begin{equation}
\label{eq:scam_residual}
\tilde{\mathbf{f}}_\ell
=
\mathbf{f}_\ell
+
\Delta\mathbf{f}_\ell,
\qquad
\Delta\mathbf{f}_\ell
=
(\boldsymbol{\gamma}_\ell-\mathbf{1})\odot\mathbf{f}_\ell+\boldsymbol{\beta}_\ell,
\end{equation}
where $\tilde{\mathbf{f}}_\ell$ are modulated CNN features.

Overall, the design adds a small modulator on top of frozen VFM and performs layer-by-layer affine operations on CNN features. Therefore, it has the following advantages: (i) computational overhead is much lower than that of heavy feature fusion; (ii) token scoring and lightweight modulation bring stability; (iii) it is plug-and-play for different CNN backbones, requiring only the acquisition of layer features and reuse of the original detection head.

\begin{table*}[thb]
\centering
\scriptsize
\renewcommand{\arraystretch}{1}
\setlength{\tabcolsep}{0.25mm}

\newcommand{\best}[1]{\textcolor{red}{\bfseries #1}}
\newcommand{\second}[1]{\textcolor{blue}{\bfseries #1}}

\begin{tabularx}{\textwidth}{|c|c|*{5}{YYYY|}}
\hline
&& \multicolumn{4}{c|}{Overall}
& \multicolumn{4}{c|}{Salient}
& \multicolumn{4}{c|}{Filamentary}
& \multicolumn{4}{c|}{Faint}
& \multicolumn{4}{c|}{Camouflaged} \\
\cline{3-22}
\multirow{-2}{*}{Scheme} & \multirow{-2}{*}{Desc.}
& IoU$\uparrow$ & nIoU$\uparrow$ & P$_d$$\uparrow$ & F$_a$$\downarrow$
& IoU$\uparrow$ & nIoU$\uparrow$ & P$_d$$\uparrow$ & F$_a$$\downarrow$
& IoU$\uparrow$ & nIoU$\uparrow$ & P$_d$$\uparrow$ & F$_a$$\downarrow$
& IoU$\uparrow$ & nIoU$\uparrow$ & P$_d$$\uparrow$ & F$_a$$\downarrow$
& IoU$\uparrow$ & nIoU$\uparrow$ & P$_d$$\uparrow$ & F$_a$$\downarrow$ \\
\hline

& Full
& 79.20 & 82.01 & 96.81 & 15.23
& 80.28 & 83.93 & 99.63 & 14.88
& 78.13 & 80.87 & 98.25 & 14.81
& 69.83 & 75.36 & 95.54 & 22.05
& 82.50 & 83.13 & 95.17 & 13.94 \\
& LESPS
& 29.12 & 40.52 & 68.57 & 34.33
& 39.75 & 57.83 & 83.21 & 31.45
& 31.23 & 41.34 & 74.75 & 35.44
& 25.32 & 43.44 & 66.07 & 38.86
& 26.07 & 29.64 & 60.14 & 36.94 \\
& PAL
& 51.11 & 59.49 & 91.43 & 30.18
& 56.22 & 69.46 & 98.13 & 33.10
& 52.30 & 59.32 & 94.25 & 33.22
& 43.21 & 61.54 & 86.61 & 29.17
& 49.91 & 53.81 & 87.86 & 33.04 \\
\cline{2-2}
\multirow{-5}{*}[-2.7ex]{\makecell{ALCLNet\\\cite{yu2022pay}}} & Ours-T
& \best{54.99} & \best{61.26} & \best{93.49} & \best{27.48}
& \best{61.61} & \best{70.57} & \best{98.88} & \second{25.25}
& \best{55.89} & \best{60.73} & \best{96.25} & \second{27.98}
& \best{47.03} & \best{62.31} & \best{91.07} & \best{23.67}
& \best{53.52} & \best{55.48} & \best{90.34} & \best{25.93} \\
& Ours-S
& \second{53.00} & \second{60.24} & \second{91.50} & \second{27.60}
& \second{59.62} & \second{70.25} & \best{98.88} & \best{25.16}
& \second{53.14} & \second{59.58} & \second{94.50} & \best{25.98}
& \second{45.42} & \second{61.65} & \second{87.50} & \second{23.85}
& \second{51.89} & \second{54.63} & \second{88.55} & \second{31.90} \\
\hline

& Full
& 80.74 & 83.63 & 96.15 & 11.03
& 83.22 & 85.49 & 100.00 & 6.69
& 77.24 & 81.74 & 96.50 & 16.26
& 74.08 & 79.10 & 93.75 & 9.20
& 83.16 & 84.85 & 94.90 & 9.71 \\
& LESPS
& 38.98 & 49.09 & 73.02 & 32.71
& 49.94 & 63.90 & 86.57 & 29.88
& 39.42 & 48.93 & 77.00 & 31.35
& 35.23 & 54.16 & 75.00 & 33.20
& 36.04 & 38.57 & 65.52 & 35.76 \\
& PAL
& 58.55 & 65.62 & 93.16 & 27.85
& 62.49 & 72.04 & 98.51 & 25.15
& 58.02 & 64.95 & 95.25 & 24.56
& 47.40 & 64.17 & 89.29 & 29.62
& 56.32 & 59.54 & 88.28 & 28.25 \\
\cline{2-2}
\multirow{-5}{*}[-2.7ex]{\makecell{GGLNet\\\cite{zhao2023gradient}}} & Ours-T
& \best{59.14} & \best{66.03} & \best{93.69} & \best{23.68}
& \best{64.07} & \best{74.34} & \best{99.25} & \second{24.29}
& \best{59.76} & \best{66.09} & \best{96.25} & \best{18.14}
& \best{48.45} & \second{64.60} & \second{91.07} & \best{28.22}
& \best{58.77} & \best{61.55} & \best{90.90} & \best{27.02} \\
& Ours-S
& \second{58.90} & \second{65.69} & \second{93.23} & \second{24.86}
& \second{63.55} & \second{73.34} & \second{98.88} & \best{24.06}
& \second{59.58} & \second{65.86} & \second{96.00} & \second{19.81}
& \second{47.95} & \best{65.45} & \best{91.96} & \second{29.55}
& \second{57.37} & \second{60.77} & \second{89.79} & \second{27.51} \\
\hline

& Full
& 79.13 & 82.76 & 95.88 & 23.37
& 79.16 & 84.96 & 99.63 & 20.21
& 77.33 & 82.49 & 97.00 & 21.26
& 69.74 & 73.30 & 90.18 & 17.70
& 80.37 & 84.66 & 94.76 & 18.34 \\
& LESPS
& 36.54 & 48.30 & 86.91 & 34.18
& 49.96 & 65.47 & 95.52 & 31.57
& 37.76 & 47.51 & 91.25 & 32.40
& 32.77 & 52.91 & 80.36 & 39.29
& 32.31 & 36.30 & 82.34 & 39.10 \\
& PAL
& 60.20 & 66.86 & 93.29 & 28.75
& 63.75 & 73.80 & 98.13 & 28.90
& 59.02 & 65.89 & 95.25 & 25.23
& 48.98 & 65.16 & 88.96 & 31.36
& 60.43 & 61.31 & 89.45 & 35.33 \\
\cline{2-2}
\multirow{-5}{*}[-2.7ex]{\makecell{MLCLNet\\\cite{yu2022infrared}}} & Ours-T
& \best{65.10} & \best{69.68} & \best{94.68} & \second{27.49}
& \best{69.80} & \best{75.78} & \second{98.51} & \second{28.68}
& \best{63.76} & \best{67.91} & \best{96.50} & \best{24.00}
& \best{56.72} & \best{69.83} & \best{92.86} & \best{17.02}
& \best{65.72} & \best{66.93} & \best{92.14} & \best{23.65} \\
& Ours-S
& \second{62.63} & \second{67.28} & \second{94.49} & \best{27.22}
& \second{68.33} & \second{74.25} & \best{99.63} & \best{27.93}
& \second{61.75} & \second{65.97} & \second{95.75} & \second{24.08}
& \second{51.54} & \second{65.54} & \second{89.61} & \second{23.60}
& \second{61.95} & \second{63.94} & \second{90.34} & \second{24.41} \\
\hline

& Full
& 80.80 & 82.67 & 96.54 & 11.73
& 81.27 & 84.76 & 99.63 & 5.34
& 80.60 & 82.04 & 97.75 & 12.73
& 70.50 & 75.64 & 89.29 & 6.77
& 81.86 & 83.23 & 95.59 & 5.63 \\
& LESPS
& 39.62 & 51.26 & 74.62 & 36.18
& 50.33 & 68.06 & 89.18 & 34.61
& 40.79 & 50.06 & 79.00 & 37.14
& 36.38 & 58.04 & 75.89 & 35.98
& 36.20 & 39.25 & 66.90 & 35.94 \\
& PAL
& 61.66 & 65.26 & 91.16 & 22.29
& 71.00 & 74.04 & 99.25 & 17.80
& 60.76 & 64.56 & 95.00 & 20.36
& 55.97 & 65.88 & 91.07 & 22.14
& 60.07 & 60.84 & 89.45 & 24.86 \\
\cline{2-2}
\multirow{-5}{*}[-2.7ex]{\makecell{MSHNet\\\cite{liu2024infrared}}} & Ours-T
& \second{62.24} & \best{66.50} & \best{94.02} & \best{18.00}
& \best{72.73} & \best{76.04} & \best{99.63} & \best{8.49}
& \second{61.27} & \best{66.31} & \best{96.50} & \second{17.97}
& \second{58.07} & \second{67.76} & \best{93.75} & \best{20.48}
& \best{61.79} & \best{62.57} & \best{90.21} & \best{23.62} \\
& Ours-S
& \best{62.36} & \second{66.26} & \second{93.36} & \second{20.52}
& \second{72.43} & \second{75.31} & \best{99.63} & \second{15.34}
& \best{61.44} & \second{65.85} & \second{95.50} & \best{14.56}
& \best{59.76} & \best{67.79} & \second{91.96} & \second{21.31}
& \second{60.46} & \second{61.44} & \second{89.52} & \second{24.44} \\
\hline

& Full
& 78.34 & 82.18 & 96.08 & 21.95
& 77.76 & 83.47 & 100.00 & 15.06
& 76.17 & 81.05 & 97.50 & 29.51
& 68.87 & 76.49 & 93.75 & 17.51
& 81.31 & 83.59 & 94.21 & 27.96 \\
& LESPS
& 35.85 & 47.63 & 72.29 & 40.76
& 47.44 & 64.34 & 88.43 & 37.83
& 37.13 & 46.89 & 76.75 & 32.53
& 33.77 & 55.25 & 73.21 & 35.89
& 31.85 & 35.62 & 63.72 & 42.60 \\
& PAL
& 58.27 & 62.28 & 92.77 & 28.62
& 63.09 & 72.22 & 99.25 & 25.76
& 60.01 & 65.39 & 96.00 & 28.51
& 43.86 & 60.69 & 84.82 & 28.38
& 56.42 & 57.58 & 86.48 & 25.06 \\
\cline{2-2}
\multirow{-5}{*}[-2.7ex]{\makecell{RepISDNet\\\cite{wu2023repisd}}} & Ours-T
& \best{64.19} & \best{69.00} & \best{95.48} & \best{24.86}
& \best{68.21} & \best{74.96} & \best{99.63} & \best{18.60}
& \best{63.74} & \best{67.57} & \best{97.00} & \second{25.63}
& \best{53.49} & \best{66.21} & \best{87.50} & \best{17.89}
& \best{63.07} & \best{63.69} & \best{91.86} & \best{15.92} \\
& Ours-S
& \second{59.47} & \second{64.56} & \second{92.96} & \second{26.99}
& \second{63.88} & \second{73.90} & \best{99.63} & \second{20.51}
& \second{60.31} & \second{65.40} & \second{96.50} & \best{24.67}
& \second{44.25} & \second{61.30} & \best{87.50} & \second{22.06}
& \second{57.59} & \second{58.82} & \second{89.52} & \second{24.62} \\
\hline

& Full
& 82.30 & 85.38 & 96.08 & 7.67
& 84.66 & 87.10 & 99.63 & 4.96
& 78.29 & 84.27 & 97.75 & 9.62
& 74.40 & 78.43 & 91.96 & 18.49
& 85.28 & 86.83 & 94.48 & 5.45 \\
& LESPS
& 38.50 & 49.41 & 71.96 & 29.07
& 51.82 & 67.29 & 88.43 & 32.65
& 38.10 & 47.07 & 75.75 & 29.03
& 32.49 & 54.44 & 72.32 & 34.76
& 35.94 & 38.37 & 63.72 & 35.95 \\
& PAL
& 62.67 & 68.22 & 92.89 & 17.59
& 66.37 & 75.05 & 98.88 & 21.75
& 62.06 & 68.34 & 96.25 & 18.55
& 49.74 & 66.67 & 90.18 & 21.71
& 62.71 & 63.84 & 88.55 & 25.39 \\
\cline{2-2}
\multirow{-5}{*}[-2.7ex]{\makecell{DNANet\\\cite{li2022dense}}} & Ours-T
& \best{63.09} & \best{69.42} & \best{95.15} & \best{16.18}
& \best{70.91} & \best{75.56} & \best{99.63} & \best{12.26}
& \best{63.74} & \best{68.67} & \second{96.50} & \second{17.85}
& \best{51.18} & \best{68.32} & \second{92.86} & \best{20.09}
& \best{63.08} & \best{66.39} & \best{93.79} & \second{22.28} \\
& Ours-S
& \second{62.81} & \second{68.54} & \second{94.62} & \second{16.68}
& \second{69.15} & \second{75.51} & \best{99.63} & \second{18.59}
& \second{63.25} & \second{68.50} & \best{97.50} & \best{16.46}
& \second{50.36} & \second{67.20} & \best{93.75} & \second{21.35}
& \second{62.77} & \second{65.19} & \second{91.31} & \best{21.56} \\
\hline

& Full
& 81.96 & 84.97 & 97.21 & 20.59
& 83.97 & 84.64 & 100.00 & 13.30
& 80.43 & 83.12 & 98.00 & 14.20
& 75.01 & 81.25 & 94.64 & 19.80
& 84.25 & 85.33 & 96.14 & 13.21 \\
& LESPS
& 37.38 & 46.78 & 75.48 & 35.18
& 49.64 & 63.36 & 85.45 & 34.91
& 39.25 & 46.56 & 78.00 & 30.84
& 35.37 & 53.05 & 75.89 & 32.73
& 32.96 & 34.86 & 70.34 & 34.07 \\
& PAL
& 65.95 & 68.57 & 93.49 & 21.24
& 72.07 & 73.09 & 98.88 & 25.37
& 66.81 & 68.67 & 97.00 & 23.06
& 57.25 & 68.31 & 90.18 & 20.45
& 64.27 & 64.88 & 89.66 & 24.54 \\
\cline{2-2}
\multirow{-5}{*}[-2.7ex]{\makecell{SCTransNet\\\cite{yuan2024sctransnet}}} & Ours-T
& \best{69.08} & \best{70.07} & \best{95.48} & \best{19.96}
& \best{75.58} & \best{75.50} & \second{99.63} & \best{18.99}
& \best{68.93} & \best{69.34} & \best{97.75} & \second{22.89}
& \best{65.94} & \best{69.32} & \best{93.75} & \best{19.09}
& \best{67.77} & \best{67.33} & \best{93.24} & \best{18.73} \\
& Ours-S
& \second{66.91} & \second{69.69} & \second{94.09} & \second{20.90}
& \second{74.72} & \second{74.72} & \best{100.00} & \second{21.29}
& \second{67.01} & \second{69.30} & \second{97.25} & \best{19.90}
& \second{61.32} & \second{68.82} & \second{91.96} & \second{20.11}
& \second{65.92} & \second{64.93} & \second{90.34} & \second{23.78} \\
\hline

\end{tabularx}
\caption{Performance comparison under different target characteristics.
We report IoU (\%), nIoU (\%), P$_d$ (\%), and F$_a$ ($10^{-6}$).}
\label{tab:characteristic_ablation_bestsecond}
\end{table*}

\subsection{Dynamic Collaborative Learning Strategy}
\label{sec:dcls}

ISTD datasets are extremely imbalanced in terms of target size, shape, SNR, and background complexity. If uniform sampling and loss averaging are adopted, the model is prone to overfitting simple samples and reducing robustness. Furthermore, dividing the training into a staged process (first evolving pseudo-labels, then distilling) fails to utilize validation feedback to correct which samples are more suitable for semantic transfer.
We propose a dynamic collaborative learning strategy, which couples (i) cluster-level reweighting with (ii) unified bilevel optimization~\cite{liu2026bilevel,liu2024task} with Gauss--Newton hypergradient. Specifically, we compute hand-crafted priors (e.g., intensity statistics, texture complexity, spectral sharpness) for each image and cluster the training and validation samples. Each cluster $c$ has a learnable logit $\alpha_c$, and the weight for each sample $i$ is obtained through softmax within the batch:

\begin{equation}
\label{eq:weight}
w_{\alpha}(x_i) =
\frac{\exp(\alpha_{c(i)})}{
\frac{1}{|\mathcal{B}|}\sum_{k\in\mathcal{B}}\exp(\alpha_{c(k)})},
\end{equation}
where $\mathcal{B}$ denotes the current mini-batch and $c(i)$ is the cluster index of sample $x_i$. Cluster-level learning avoids overfitting to a small number of simple samples and promotes balanced improvement across different levels of scenarios.

To enhance the efficiency of bilevel learning, we adopt a Gauss--Newton style hypergradient approximation based on gradient alignment.
Specifically, the hypergradient for $\alpha$ is corrected by an implicit term induced by the inner update:
\begin{equation}
\label{eq:gn_coeff}
\begin{aligned}
\tilde{\nabla}_{\alpha}L_{\mathrm{out}}
&\approx
\nabla_{\alpha}L_{\mathrm{out}}
+
\eta\,\rho\,\nabla_{\alpha}L_{\mathrm{in}}, \\
\text{s.t.}\quad
\rho&=\frac{\left\langle \nabla_{\theta}L_{\mathrm{out}},\,\nabla_{\theta}L_{\mathrm{in}}\right\rangle}
{\left\|\nabla_{\theta}L_{\mathrm{in}}\right\|_2^2+\epsilon},
\end{aligned}
\end{equation}
where $\eta$ is the inner-step learning rate, $\langle\cdot,\cdot\rangle$ denotes the inner product, and $\epsilon$ is a small constant for numerical stability.
This Gauss--Newton correction avoids explicit Hessian inversion and maintains stability as the pseudo-mask continuously evolves. The whole algorithm is summarized in Alg.~\ref{alg:bilevel}.

\begin{algorithm}[t]
\caption{Dynamic Collaborative Learning Strategy}
\label{alg:bilevel}
\small
\begin{algorithmic}
\Statex \textbf{Input:}
\Statex \quad Training set $\mathcal{D}_{\mathrm{tr}}$, validation set $\mathcal{D}_{\mathrm{val}}$, frozen VFM;
\Statex \quad student parameters $\theta$, SCAM parameters $\phi$, cluster logits $\alpha$
\Statex \textbf{Initialize:} pseudo-masks $\tilde{y}$ from point annotations
\For{each epoch}
    \State Update pseudo-masks $\tilde{y}$ using teacher prediction $p_t$
    \State Sample mini-batch $\mathcal{B}_{\mathrm{tr}} \sim \mathcal{D}_{\mathrm{tr}}$ and compute weights $w_{\alpha}$
    \State \textbf{Inner step:} $\phi \leftarrow \phi - \eta \nabla_{\phi} L_{\mathrm{in}}(\mathcal{B}_{\mathrm{tr}})$
    \State Sample mini-batch $\mathcal{B}_{\mathrm{val}} \sim \mathcal{D}_{\mathrm{val}}$ and compute weights $w_{\alpha}$
    \State \textbf{Outer step:} update $\theta,\alpha$ by $\nabla_{\theta,\alpha} L_{\mathrm{out}}(\mathcal{B}_{\mathrm{val}})$ using Eq.~\eqref{eq:gn_coeff}
\EndFor
\end{algorithmic}
\end{algorithm}

\section{Experiments}

\subsection{Implementation Configurations}

\paragraph{Datasets.}
The comprehensive dataset SIRST3~\cite{Ying_2023_CVPR} consists of three sub-datasets SIRST-v1~\cite{dai2021asymmetric}, NUDT-SIRST~\cite{li2022dense}, and IRSTD-1k~\cite{zhang2022isnet}. 
Following the LESPS split protocol, all models are trained on SIRST3, with 10\% of the samples held out for validation (1402/274/1079 for train/val/test).

We further report results on a test split into Salient, Filamentary, Faint, and Camouflaged, defined by target shape difficulty and target-to-background contrast. Specifically, Salient indicates regular and high-contrast targets, Filamentary indicates morphology-challenging but high-contrast targets, Faint indicates low-contrast but less complex targets, and Camouflaged represents the hardest cases with both challenging morphology and low contrast. This split enables a more interpretable, challenge-aware comparison.

\paragraph{Experimental settings.}

Unless otherwise specified, we train for 300 epochs with AdamW (batch size 16). The base learning rate is $1\times10^{-3}$ for both inner and outer updates; for backbones that are less stable under weak supervision, we use $5\times10^{-4}$. We adopt DINOv3~\cite{simeoni2025dinov3} ViT-S+/16 pretrained on LVD-1689M and keep the VFM backbone frozen throughout. We use PAL~\cite{yu2025easy} as the default pseudo-mask evolution engine, while LESPS~\cite{Ying_2023_CVPR} is reported as a baseline. Bilevel updates are performed every 5 epochs: we sample mini-batches from the training and validation sets and run 4 Gauss--Newton hypergradient steps to update the ISTD CNN, SCAM parameters, and cluster weights $\alpha$. We report IoU and nIoU for mask quality, P$_d$ for detection probability (recall), and F$_a$ for false alarm rate, following standard ISTD evaluation.
 
\subsection{Experimental Results}

\paragraph{Quantitative results.}
Table~\ref{tab:characteristic_ablation_bestsecond} reports results on SIRST3 over the overall test set and four characteristic partitions, including Salient, Filamentary, Faint, and Camouflaged, across seven backbones.
Compared with single-point baselines such as LESPS and PAL, our method achieves consistent improvements across metrics, while the deployable student (Ours-S) remains close to the VFM-embedded teacher (Ours-T).
PAL is a stronger baseline than LESPS but still suffers from missed targets and higher false alarms, especially on harder subsets, whereas our gains are most evident on Faint and Camouflaged samples, indicating improved robustness under distribution shift.
For example, on DNANet, Ours-T improves overall IoU from 62.67 to 63.09 and reduces F$_a$ from 17.59 to 16.18; on SCTransNet, Ours-T further boosts IoU from 65.95 to 69.08 with lower F$_a$, and achieves a large gain on {Faint} targets. 
Consistent improvements are observed across additional backbones, including ALCLNet, GGLNet, and MLCLNet, where both Ours-T and Ours-S improve IoU/nIoU and P$_d$ while keeping F$_a$ competitive.
Overall, these results validate that our bilevel, generalization-driven knowledge distillation effectively transfers VFM priors into a lightweight CNN student for robust detection across diverse target characteristics.

\paragraph{Qualitative results.}

\begin{figure*}[!htb]
	\centering
	\setlength{\tabcolsep}{1pt} 
	\begin{tabular}{c}		
		\includegraphics[width=0.99\textwidth]{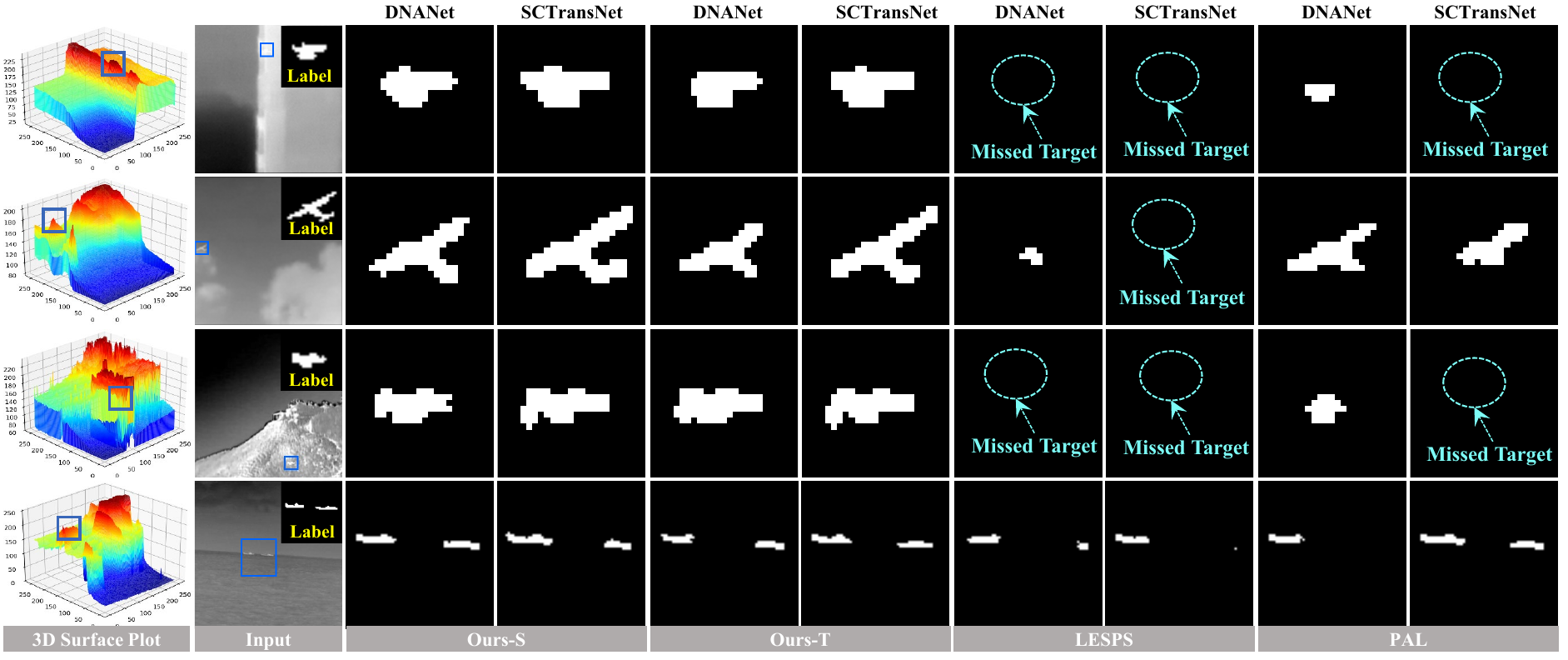}\\
	\end{tabular}
	\caption{
Qualitative comparison on challenging infrared small-target scenarios.
We visualize predicted masks from our student (Ours-S) and VFM-embedded teacher (Ours-T) instantiated on DNANet / SCTransNet, and compare them with LESPS and PAL.
}
\label{fig:Qualitative}
\end{figure*}

Fig.~\ref{fig:Qualitative} presents qualitative comparisons produced by different methods under two backbones, DNANet and SCTransNet.
Our approach yields consistently more complete and compact target responses with clearer boundaries in both the teacher and the student. 
Notably, even under extremely low contrast, tiny target size, or heavy background clutter, Ours-S remains capable of activating the true target region while effectively suppressing spurious background responses. 
In contrast, LESPS and PAL frequently suffer from missed detections, which are marked as Missed Target, or fragmented/shifted predictions across multiple cases. 
Despite discarding the VFM branch at inference, Ours-S remains close to Ours-T, indicating that our bilevel knowledge distillation effectively transfers VFM generalizable semantics into the lightweight student for robust ISTD.

\subsection{Ablation Studies}

\paragraph{Training Stability Analysis.}

\begin{figure}[t]
        	\centering
        	\includegraphics[width=0.98\columnwidth]{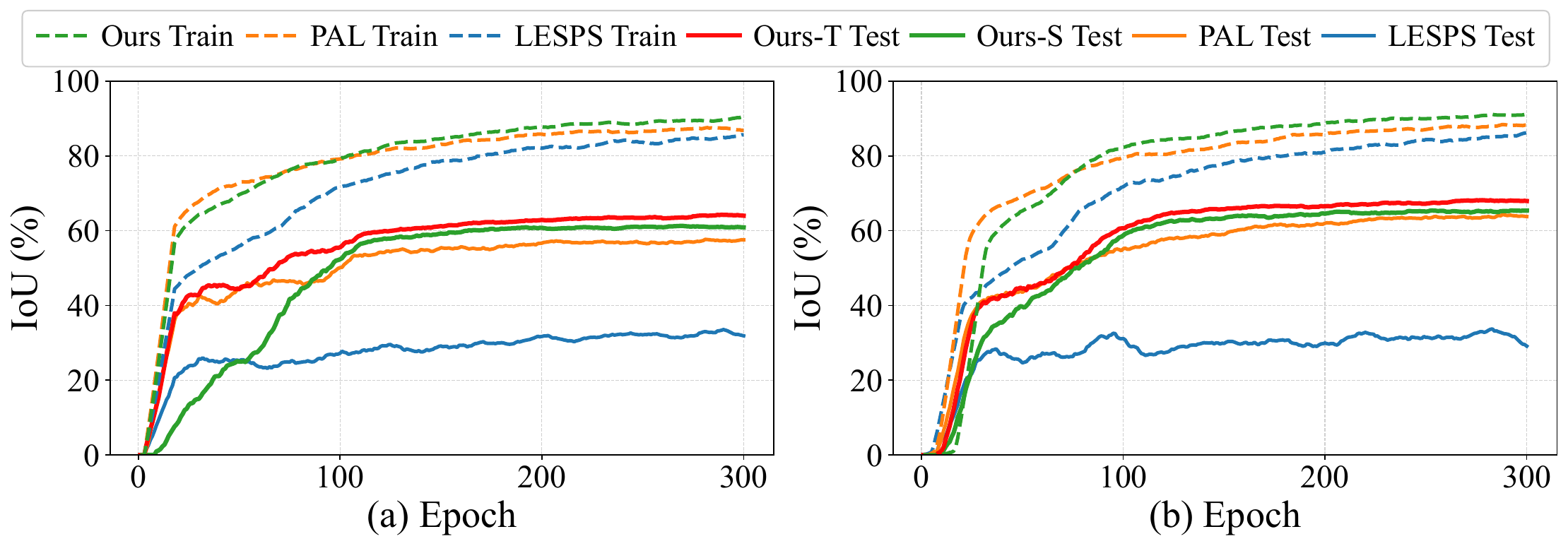} 
        	\caption{
Training stability analysis.
IoU curves on the training and test sets for MLCLNet (a) and SCTransNet (b).
}
\label{fig:stable}
\end{figure}

ISTD is challenging due to extremely small targets and scarce annotations. Under point-supervised label evolution, the pseudo-mask $\tilde{y}$ is noisy and constantly changing, which often amplifies training instability and overfitting.
Our VFM-embedded teacher injects robust global semantics via SCAM, while the bilevel optimization leans more towards transferable cues. Fig.~\ref{fig:stable} shows higher final IoU on both train and test, with smoother optimization and reduced overfitting.
Overall, these observations validate our design goals: enhancing VFM-driven optimization, stabilizing training under incomplete supervision conditions, and distilling this stability into an ISTD CNN student.

\paragraph{Impact of VFM on CNN Representations.}

\begin{figure*}[t]
	\centering
	\setlength{\tabcolsep}{1pt} 
	\begin{tabular}{c}		
		\includegraphics[width=0.99\textwidth]{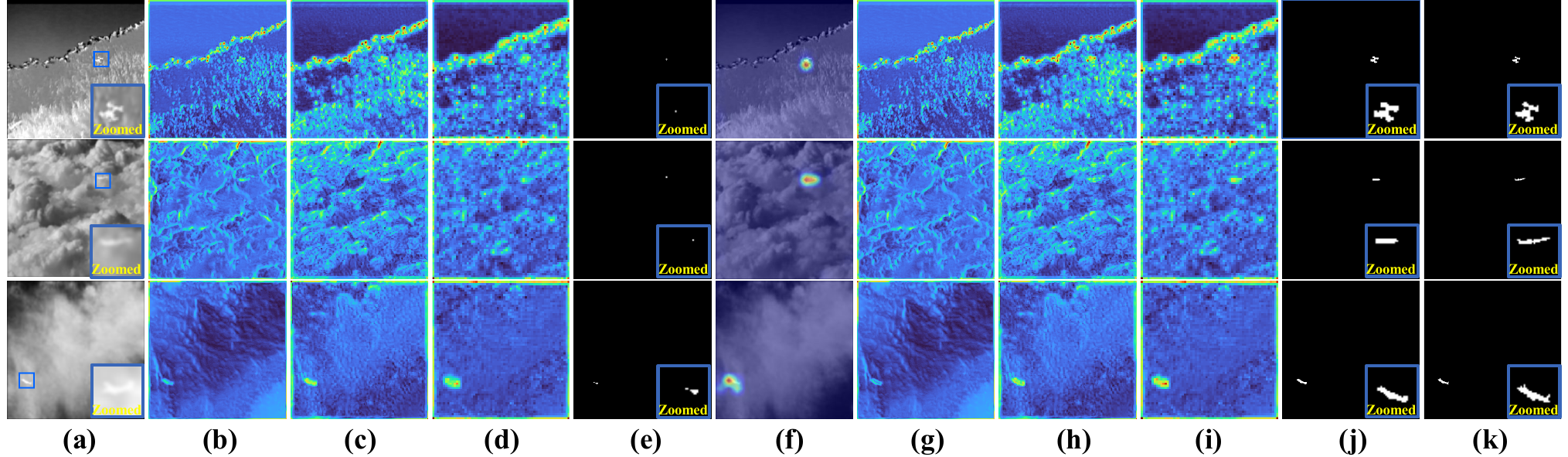}\\
	\end{tabular}
	\caption{
Impact of VFM semantics on CNN representations.
(a) Input infrared image.
(b--d) Encoder features from three layers before SCAM modulation.
(e) Prediction of the unmodulated network.
(f) DINO attention (score) map indicating semantic regions of interest.
(g--i) Corresponding layer features after SCAM modulation.
(j) Prediction of the modulated network.
(k) Ground-truth label.
}
\label{fig:dino_impact}
\end{figure*}

To visually demonstrate how VFM priors reshape ISTD detectors, we visualize the intermediate representations of the CNN encoder before and after SCAM modulation. Specifically, for three encoder layers, we calculate the channel mean intensity maps of the original features ${\mathbf{f}_\ell}$ and modulated features ${\tilde{\mathbf{f}}_\ell}$. Additionally, we also visualize the VFM patch grid score map generated by token-aware aggregation, which directly reflects the attention distribution of DINO patch tokens.
Fig.~\ref{fig:dino_impact} shows SCAM boosts target responses and suppresses clutter across layers, producing clearer spatial contrast.
The VFM score maps often focus on subtle target cues that are easily overlooked by purely local CNN priors (e.g., faint or irregular targets). Therefore, VFM can provide complementary global guidance. This effect ultimately translates into more accurate and compact predicted shapes in difficult samples and improves overall detection performance.

\paragraph{Computational Overhead.}

\begin{table}[t]
\centering
\scriptsize
\setlength{\tabcolsep}{4pt}
\renewcommand{\arraystretch}{1}
\resizebox{\columnwidth}{!}{%
\begin{tabular}{c c|cc cc|cc}
\toprule
\multirow{2}{*}[-0.6ex]{Backbone} & \multirow{2}{*}[-0.6ex]{Params (M)}
& \multicolumn{2}{c}{Student}
& \multicolumn{2}{c}{Teacher}
& \multicolumn{2}{c}{Rel. Overhead} \\
\cmidrule(lr){3-4}\cmidrule(lr){5-6}\cmidrule(lr){7-8}
& 
& Lat.$\downarrow$ & Mem$\downarrow$
& Lat.$\downarrow$ & Mem$\downarrow$
& Lat.\,$(\times)\downarrow$ & Mem\,$(\times)\downarrow$ \\
\midrule
MLCLNet    & 0.66  & 34.54  & 939.6  & 54.22  & 1051.8 & 1.57 & 1.12 \\
DNANet     & 4.70  & 529.25 & 1263.0 & 548.39 & 1387.3 & 1.04 & 1.10 \\
ALCLNet    & 5.67  & 30.47  & 555.5  & 49.63  & 681.4  & 1.63 & 1.23 \\
SCTransNet & 11.33 & 223.40 & 1314.1 & 265.33 & 1742.6 & 1.19 & 1.33 \\
\bottomrule
\end{tabular}}
\caption{Inference overhead comparison across four backbones.
We report the mean latency (ms) and peak GPU memory (MB).}
\label{tab:eff_overhead}
\end{table}

We benchmark our framework on four backbones using a single NVIDIA A40 GPU with AMP enabled. We perform $20$ warm-up iterations and report averaged latency over $100$ measured iterations, repeated $3$ times.
The student is unchanged for inference, while the teacher is equipped with a frozen DINOv3 ViT-S+/16 backbone with about 29M parameters.
As summarized in Table~\ref{tab:eff_overhead}, the teacher introduces a moderate additional cost with a latency increase ranging from $1.04{\times}$ to $1.63{\times}$ and a peak memory increase from $1.10{\times}$ to $1.33{\times}$, depending on the backbone.
For SCTransNet, overhead adds $1.19{\times}$ latency and $1.33{\times}$ memory. On lightweight backbones (e.g., MLCLNet / ALCLNet), latency overhead becomes more visible but memory remains efficient.
Importantly, since all teacher components are removed for deployment, our improvements come from the proposed training strategy without increasing inference-time complexity of the student model.

\paragraph{Efficiency of Proposed Mechanisms.}

\begin{figure}[t]
        	\centering
        	\includegraphics[width=0.98\columnwidth]{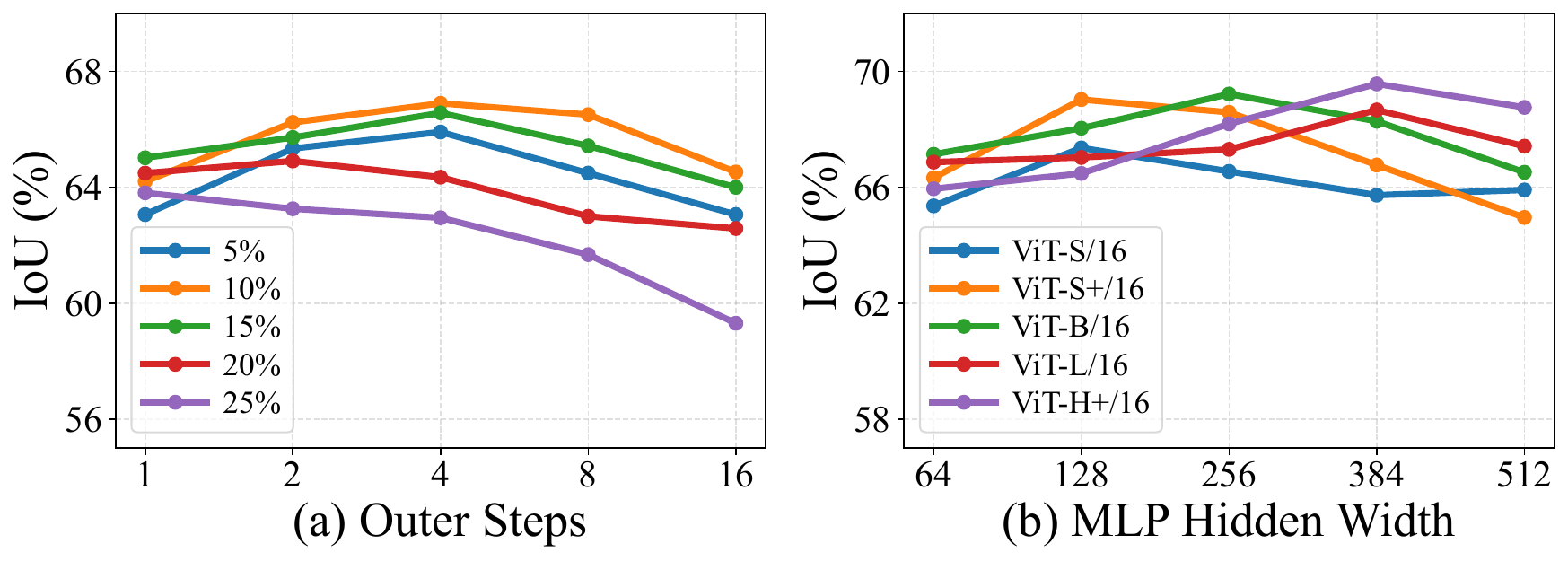} 
        	\caption{
(a) IoU versus the number of outer optimization steps under different validation ratios (Student).
(b) IoU versus the hidden width of the SCAM MLP for different DINOv3 variants (Teacher).
}
\label{fig:21}
\end{figure}

Despite adding a frozen VFM branch and hierarchical training, the gains come from a few lightweight designs. Table~\ref{tab:overall_checklist} shows that removing VFM and SCAM decreases IoU to 63.06, suggesting pure CNNs lack semantic discrimination with evolving pseudo-masks.
The two knowledge distillation terms are complementary: removing either lowers IoU and raises F$_a$. The inner term $L_{\mathrm{kd}}(p_t,p_s)$ stabilizes teacher adaptation, while the outer term $L_{\mathrm{kd}}(p_s,p_t)$ distills validation semantics from $\mathcal{D}_{\mathrm{val}}$ into the student.
Removing the validation set causes one of the largest performance drops, showing validation feedback counters cross-scene bias; while disabling cluster-level reweighting reduces P$_d$ and increases F$_a$, indicating it mitigates imbalance.
Fig.~\ref{fig:21} (a) shows performance saturates with 10\% $\mathcal{D}_{\mathrm{val}}$ and 4 outer steps; Fig.\ref{fig:21} (b) indicates that the hidden layer width of the MLP in SCAM depends on the DINOv3 variant due to different semantic dimensions, so the overhead is controllable without sensitive tuning.

\begin{table}[t]
\centering
\scriptsize
\setlength{\tabcolsep}{2pt}
\renewcommand{\arraystretch}{1}
\resizebox{\columnwidth}{!}{%
\begin{tabular}{lcccccc|cccc}
\toprule
Method & VFM & SCAM & Outer-KD & Inner-KD & Val Set & Reweight
& IoU$\uparrow$ & nIoU$\uparrow$ & P$_d$$\uparrow$ & F$_a$$\downarrow$ \\
\midrule
w/o VFM              & \xmark & \xmark & \xmark & \xmark & \xmark & \xmark & 63.06 & 64.77 & 93.49 & 29.27 \\
w/o $L_{\mathrm{kd}}(p_s,p_t)$ & \cmark & \cmark & \xmark & \cmark & \cmark & \cmark & 64.35 & 68.75 & 93.09 & 32.80 \\
w/o $L_{\mathrm{kd}}(p_t,p_s)$ & \cmark & \cmark & \cmark & \xmark & \cmark & \cmark & 64.45 & 67.38 & 93.42 & 26.99 \\
w/o $\mathcal{D}_{\mathrm{val}}$        & \cmark & \cmark & \cmark & \cmark & \xmark & \cmark & 63.33 & 66.07 & \textcolor{red}{\textbf{95.22}} & 22.43 \\
$w_{\alpha}(x)\equiv1$          & \cmark & \cmark & \cmark & \cmark & \cmark & \xmark & 65.22 & 67.31 & 91.30 & 23.72 \\
\midrule
Proposed & \cmark & \cmark & \cmark & \cmark & \cmark & \cmark & \textcolor{red}{\textbf{66.91}} & \textcolor{red}{\textbf{69.69}} & 94.09 & \textcolor{red}{\textbf{20.90}} \\
\bottomrule
\end{tabular}}
\caption{Ablation study of the proposed core components.}
\label{tab:overall_checklist}
\end{table}

\paragraph{SCAM Design.}

\begin{table}[t]
\centering
\scriptsize
\setlength{\tabcolsep}{4pt}
\renewcommand{\arraystretch}{1}
\resizebox{\columnwidth}{!}{%
\begin{tabular}{l|cccc cccc}
\toprule
\multirow{2}{*}[-0.6ex]{Method} & \multicolumn{4}{c}{Teacher} & \multicolumn{4}{c}{Student} \\
\cmidrule(lr){2-5}\cmidrule(lr){6-9}
& IoU$\uparrow$ & nIoU$\uparrow$ & P$_d$$\uparrow$ & F$_a$$\downarrow$ & IoU$\uparrow$ & nIoU$\uparrow$ & P$_d$$\uparrow$ & F$_a$$\downarrow$ \\
\midrule
Naive Concat Fusion & 64.99 & 69.58 & 94.95 & 25.32 & 64.09 & 69.17 & 93.36 & 28.00 \\
w/o TAP (GAP instead) & 63.31 & 62.43 & 94.15 & 30.43 & 63.09 & 63.60 & 94.15 & 33.95 \\
Single-depth VFM Feature & 64.40 & 68.37 & 93.89 & 32.41 & 63.27 & 66.42 & 94.42 & 27.72 \\
Single-layer Modulation & 66.43 & 69.77 & \textcolor{red}{\textbf{95.55}} & 27.72 & 64.76 & 69.49 & \textcolor{red}{\textbf{94.55}} & 23.03 \\
w/o $\mathcal{R}_{\mathrm{gate}}$ & 68.36 & 69.90 & 94.62 & \textcolor{red}{\textbf{18.54}} & 66.18 & 69.23 & 93.69 & \textcolor{red}{\textbf{17.85}} \\
\midrule
Proposed & \textcolor{red}{\textbf{69.08}} & \textcolor{red}{\textbf{70.07}} & 95.48 & 19.96 & \textcolor{red}{\textbf{66.91}} & \textcolor{red}{\textbf{69.69}} & 94.09 & 20.90 \\
\bottomrule
\end{tabular}}
\caption{Ablation study of SCAM.}
\label{tab:scam_ablation}
\end{table}

\begin{table}[t]
\centering
\scriptsize
\setlength{\tabcolsep}{8pt}
\renewcommand{\arraystretch}{1}
\resizebox{\columnwidth}{!}{%
\begin{tabular}{l|cc cc cc}
\toprule
\multirow{2}{*}[-0.6ex]{Subset}
& \multicolumn{2}{c}{$H_{\mathrm{norm}}\downarrow$}
& \multicolumn{2}{c}{$\mathrm{EffN}$$\downarrow$}
& \multicolumn{2}{c}{$p_{\max}\uparrow$} \\
\cmidrule(lr){2-3}\cmidrule(lr){4-5}\cmidrule(lr){6-7}
& Mean & Std & Mean & Std & Mean & Std \\
\midrule
Salient
& 81.73 & 9.91 & 58.17 & 34.55 & 9.96 & 8.65 \\
Filamentary
& 79.83 & 11.60 & 48.85 & 31.97 & 11.80 & 11.34 \\
Faint
& 81.67 & 9.74 & 59.20 & 33.75 & 10.43 & 9.15 \\
Camouflaged
& \textbf{69.36} & 12.42 & \textbf{32.05} & 24.29 & \textbf{17.59} & 15.60 \\
\bottomrule
\end{tabular}}
\caption{Characteristic-wise attention analysis of DINO modulator.
$H_{\mathrm{norm}}$ (\%) denotes the normalized attention entropy,
$\mathrm{EffN}$ denotes the effective number of attended patches, and $p_{\max}$ (\%) denotes the maximum attention weight.}
\label{tab:dino_cluster_attn_split}
\end{table}

In Table~\ref{tab:scam_ablation}, we compared six VFM-CNN fusion strategies to better integrate VFM priors with ISTD CNN detectors. Replacing TAP with Global Average Pooling (GAP) or using a single VFM depth degrades performance; GAP increases F$_a$ to 33.95. This suggests that selective token aggregation and multi-depth fusion are crucial in cluttered infrared backgrounds.
For semantic injection, naive concatenation or single-layer modulation is inferior, while multi-layer affine modulation brings stronger and more stable gains, better aligning global semantics with fine-grained localization. 
Removing the layer-wise gate reduces IoU to 66.18 and P$_d$ to 93.69, implying that gated semantic control is important under evolving pseudo-masks $\tilde{y}$.
Table~\ref{tab:dino_cluster_attn_split} shows that camouflaged targets have lower $H_{\mathrm{norm}}$, smaller $\mathrm{EffN}$, and larger $p_{\max}$, i.e., attention becomes more selective under heavy clutter and irregular shapes.
Overall, SCAM balances performance and robustness via token-aware multi-depth aggregation, multi-layer affine modulation, and gated injection, enabling layer-wise guidance for stable bilevel transfer.

\section{Conclusion}

This paper proposes a bilevel knowledge distillation framework guided by a VFM for point-supervised ISTD. During training, the VFM is frozen and its knowledge is distilled into a lightweight detector. SCAM and dynamic collaborative learning reduce the impact of noisy pseudo-masks and stabilize training. Experiments on SIRST3 show consistent improvements across different backbones and data splits.

\appendix

\section*{Acknowledgments}

This work is partially supported by  the National Natural
Science Foundation of China (Nos. 62306059, 624B2033),
the Fundamental Research Funds for the Central Universities.

\bibliographystyle{named}
\bibliography{ijcai26}

\end{document}